\documentclass[11pt,a4paper]{article}
\usepackage[hyperref]{eacl2021}
\usepackage{times}
\usepackage{latexsym}

\usepackage{microtype}

\aclfinalcopy

\usepackage{adjustbox}
\usepackage{booktabs}
\usepackage{xspace}
\usepackage{amssymb}
\usepackage{interval}
\usepackage{multirow}
\usepackage{multicol}
\usepackage{supertabular}

\graphicspath{{./fig/}{./fig_raiden/}}

\newcommand{\fig}{Fig.~}
\newcommand{\appx}{Appx.~}
\newcommand{\ex}[1]{``#1''}
\newcommand{\term}{\textbf}
\newcommand{\mask}{[MASK]}
\newcommand{\entity}[1]{{\small{\textsf{#1}}}}
\newcommand{\predicate}{\entity}
\newcommand{\triple}[3]{$\langle$\entity{#1}, \predicate{#2}, \entity{#3}$\rangle$}
\newcommand{\inparatitle}[1]{\noindent\textbf{#1}.}
\newcommand{\inparatitlec}[1]{\noindent\textbf{#1}:}
\newcommand{\researchquestion}[1]{\textbf{#1}}
\newcommand{\secref}[1]{(\S\ref{#1})}

\title{%
Language Models as Knowledge Bases:\\
On Entity Representations, Storage Capacity, and Paraphrased Queries}

\author{%
Benjamin Heinzerling\textsuperscript{\textnormal{1, 2}} \textnormal{and} Kentaro Inui\textsuperscript{\textnormal{2, 1}} \\
\textsuperscript{\textnormal{1}}RIKEN AIP \& \textsuperscript{\textnormal{2}}Tohoku University \\
\hypersetup{urlcolor=black} \href{mailto:benjamin.heinzerling@riken.jp}{\tt benjamin.heinzerling@riken.jp} \hspace{0.06em} $\vert$ \hspace{0.12em} \href{mailto:inui@tohoku.ac.jp}{\tt inui@tohoku.ac.jp}%
}
\date{}

\begin{document}
\maketitle
\begin{abstract}
Pretrained language models have been suggested as a possible alternative or complement to structured knowledge bases.
However, this emerging LM-as-KB paradigm has so far only been considered in a very limited setting, which only allows handling 21k entities whose name is found in common LM vocabularies.
Furthermore, a major benefit of this paradigm, i.e., querying the KB using natural language paraphrases, is underexplored.
	Here we formulate two basic requirements for treating LMs as KBs: (i) the ability to store a large number facts involving a large number of entities and (ii) the ability to query stored facts. 
We explore three entity representations that allow LMs to handle millions of entities and present a detailed case study on paraphrased querying of facts stored in LMs, thereby providing a proof-of-concept that language models can indeed serve as knowledge bases.
\end{abstract}

\section{Introduction}

\begin{figure}[t!]
	\centering
	\includegraphics[width=\linewidth]{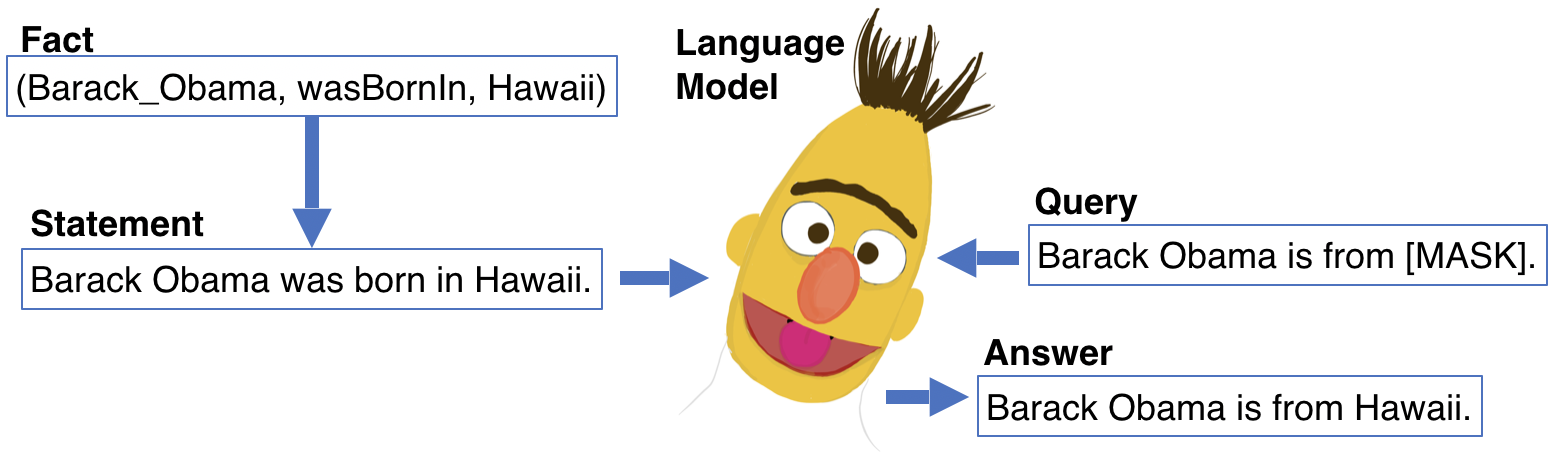}
	\vspace{-2ex}
	\caption{The LM-as-KB paradigm, introduced by \citet{petroni2019language}. A LM memorizes factual statements, which can be queried in natural language.}
	\label{fig:lm-as-kb}
\end{figure}

Language models (LMs) appear to memorize world knowledge facts during training.
For example, BERT \citep{devlin2019bert} correctly answers the query \ex{Paris is the capital of \mask} with \ex{France}.
This observation prompted \citet{petroni2019language} to ask if LMs can serve as an alternative or complement to structured knowledge bases (KBs), thereby introducing the idea of treating LMs as KBs: During training, the LM encounters world knowledge facts expressed in its training data, some of which are stored in some form in the LM's parameters.
After training, some of the stored facts can be recovered from the LM's parameters by means of a suitable natural language query (\fig\ref{fig:lm-as-kb}).
A LM with such a ``built-in'' KB is useful for knowledge-intensive tasks \cite{petroni2020kilt} and question answering \cite{roberts2020knowledge}, and could improve natural language interfaces to structured data \cite{hendrix1978natural,herzig2020tapas}.
However, this emerging LM-as-KB paradigm is faced with several foundational questions.

\inparatitlec{First question}
KBs contain millions of entities, while LM vocabulary size usually does not exceed 100k entries. \researchquestion{How can millions of entities be represented in LMs?}
\citet{petroni2019language} circumvent this problem by only considering 21k entities whose canonical name corresponds to a single token in the LM's vocabulary, e.g., entities like \ex{France} or \ex{Bert}, but not \ex{United Kingdom} or \ex{Sesame Street}.
Hence, this approach cannot handle entities not contained in the vocabulary and a query like \ex{Bert is a character on \mask} is not answerable in this simplified setting.
To answer this first question, we compare three methods for scaling LM-as-KB to millions of entities: %
\begin{enumerate}
	\item Symbolic representation, i.e., extending the LM vocabulary with entries for all entities;
	\item Surface form representation, i.e., each entity is represented by its subword-encoded canonical name, which is stored and queried by extending the LM with a sequence decoder; and
	\item Continuous representation, i.e., each entity is represented as an embedding.
\end{enumerate} %
We find that, while all three entity representations allow LMs to store millions of world-knowledge facts involving a large number of entities, each representation comes with different trade-offs:
Symbolic representation allows the most accurate storage, but is computationally expensive and requires entity-linked training data.
Surface representation is computationally efficient and does not require entity-linked data, but is less accurate, especially for longer entity names.
Continuous representation also requires entity-linked data, but is computationally more efficient than symbolic representation.

\inparatitlec{Second question}
\researchquestion{What is the capacity of LMs for storing world knowledge?} Can a LM store, say, all relation triples contained in a KB like Wikidata \citep{vrandevcic2014wikidata}?
Here we conduct experiments using synthetic data to study the scaling behaviour of current LM architectures.
Varying the number of trainable model parameters and recording the number of relation triples memorized at a given accuracy level, we find that, e.g., a Transformer \cite{vaswani2017attention} with 125 million parameters (12 layers of size 768),
has the capacity to memorize 1 million Wikidata relation triples with 95 percent accuracy or 5 million relation triples with 79 percent accuracy.
Assuming linear scaling, this finding suggests that larger LMs with tens or hundreds of billions of parameters \citep{raffel2019exploring,brown2020language} can be used to store sizable parts, if not all, of a KB like Wikidata.

\inparatitlec{Third question}
\researchquestion{How robustly is world knowledge stored in LMs?} Is the LM able to recall a fact even if the query is slightly different than what was memorized during training?
For example, if the LM memorized \ex{Barack Obama was born in Hawaii}\, during training, can it answer queries like \ex{Barack Obama is from \mask}\, or \ex{Where was Barack Obama born? \mask}?
Here we conduct experiments to measure how well the LM transfers knowledge from memorized statements to query variants, both in a zero-shot setting in which the model is not exposed to the target query variant during training, and a few shot setting, in which the model is finetuned on a small number of statements containing the target query variant.
We observe zero-shot transfer in case of highly similar query variants, and see successful few-shot transfer after finetuning with 5 to 100 instances in case of less similar queries.
This ability to handle soft, natural language queries, as opposed to hard, symbolic queries in a language like SQL or SPARQL, is one of the key motivations for using LMs as KBs.

\inparatitle{Contributions} 
We formulate two requirements for treating LMs as KBs: (i) the ability to store a large number of facts involving a large number of entities and (ii) the ability to query stored facts.
After providing background on world knowledge in LMs \secref{sec:background}, we make the following contributions:\footnote{%
	Code available at:\\
	\url{https://github.com/bheinzerling/lm-as-kb}
}%
\begin{itemize}%
	\item A comparison of entity representations for scaling LM-as-KB to millions of entities \secref{sec:entity-representations};
	\item Empirical lower bounds on LM capacity for storing world knowledge facts \secref{sec:capacity}; and
	\item A controlled study of knowledge transfer from stored facts to paraphrased queries \secref{sec:query}.
\end{itemize}%
\inparatitle{Terminology} In this work we are interested in storing and retrieving world knowledge facts in and from a LM.
\term{World knowledge} is knowledge pertaining to entities, such as \entity{Barack\_Obama}.
A \term{fact} is a piece of world knowledge that can be expressed with a concise natural language \term{statement}, such as the English sentence \emph{Barack Obama was Born in Hawaii}, or with a \term{relation triple}, such as \triple{Barack\_Obama}{wasBornIn}{Hawaii}.
A relation triple, or relation for short, consists of a \term{subject} entity (\entity{Barack\_Obama}), a \term{predicate} (\predicate{wasBornIn}), and an \term{object} entity (\entity{Hawaii}).
A \term{knowledge base} is a set of relations.
Knowledge bases, such as Wikidata, typically contain thousands of predicates, millions of entities, and billions of relations.

\section{World Knowledge in Language Models}
\label{sec:background}

Large pretrained LMs have been the driver of recent progress in natural language processing \citep{peters2018deep,howard2018universal,radford2019language,devlin2019bert}.
While the trend towards larger LMs is likely to continue \citep{raffel2019exploring,kaplan2020scaling,brown2020language}, it has limitations:
(i) A model trained only on text lacks grounding in perception and experience and hence cannot learn meaning \cite{bender2020climbing}.
(ii) Reporting bias leads to certain knowledge rarely or never being expressed in text.
For example, a LM will easily learn to associate the phrase ``Barack Obama'' with the phrase ``U.S.\, President'', but might less likely learn that he is a ``human being'', since the latter fact is rarely stated explicitly in text.
In contrast, this type of knowledge is readily available in KBs.
(iii) A large number of rare entities \citep{hoffart2014emerging,derczynski2017wnut,ilievski2018long} are, by definition, rarely mentioned, making it difficult for LMs to acquire knowledge about this long tail of entities from text alone.

These limitations have motivated efforts to explicitly\footnote{As opposed to the LM acquiring world knowledge implicitly as a side effect of its training objective.} equip LMs with world knowledge.
Table~\ref{tbl:rel-work} (\appx\ref{sec:relwork-tbl}) situates these efforts on a spectrum from purely text-based LMs to representations of structured KBs.
Models based on text generation \citep{raffel2019exploring,roberts2020knowledge} and retrieval \citep{guu2020realm} have proven most successful in knowledge-intensive tasks.
However, we argue that models which reify entities \citep{logan2019baracks}, i.e., models in which entities are ``first-class citizens'' that can be directly predicted\footnote{As opposed to generating or retrieving a surface form which may or may not correspond to an entity.}, are a promising research direction, since the direct links into a KB can be seen as a form of grounding.
This is one of our main motivations for considering symbolic and continuous entity representations.

\section{Entity Representations}
\label{sec:entity-representations}

How can millions of entities be represented in a LM? To answer our first question, we compare three types of entity representations: symbolic, surface form, and continuous.

\inparatitle{Experimental setup} We evaluate entity representations by measuring how well they allow a LM to store and retrieve world knowledge facts.
For example, if the LM's training data contains the statement \ex{Bert is a character on Sesame Street}, the model should memorize this statement and recall the correct object \entity{Sesame\_Street} when asked with a query like \ex{Bert is a character on \mask.}

\inparatitle{Synthetic data}
It is not a priori clear how many facts a text from the LM's training data, say, a Wikipedia article, expresses.
Since we want to precisely measure how well a LM can store and retrieve facts, we create synthetic data by generating statements from KB relations and then train the model to memorize these statements.
Using Wikidata as KB, we first define two sets of entities:
A smaller set consisting of the top 1 million Wikidata entities according to node outdegree, and a larger set consisting of the roughly 6 million Wikidata entities that have an entry in the English Wikipedia.

Next, we manually create templates for the 100 most frequent Wikidata predicates.
For example, for the predicate \predicate{P19} (\ex{place of birth}), we create the template \emph{S was born in O} and generate English statements by filling the \emph{S} and \emph{O} slots with entities from the sets defined above for which this relation holds.%
\footnote{Templates and sample of statements in \appx\ref{sec:templates} and \ref{sec:statements}.}
To make queries for an object unique given subject and predicate, we arbitrarily select exactly one fact if there are multiple objects and discard the other facts.
This process yields 5 million statements involving up to 1 million entities, and 10 million statements involving up to 6 million entities.
These statements then serve as training instances, i.e., given the query \ex{Barack Obama was born in \mask}, the model should predict \entity{Hawaii}.
As our goal is to store facts in a LM, there is no distinction between training and test data.

\inparatitle{Models and training}
We consider two common LM architectures: LSTMs \citep{hochreiter1997lstm} and Transformers \citep{vaswani2017attention}.
For LSTMs, we compare two configurations; a randomly initialized two-layer LSTM with layers size 256 (\emph{LSTM 256}) and one with layer size 1024 (\emph{LSTM 1024}).
For Transformers, we compare a pretrained RoBERTa-base \citep{liu2019roberta}, and RoBERTa without pretraining, i.e., a randomly initialized Transformer of the same size.
For consistent tokenization across all four models, we subword-tokenize statements with the RoBERTa tokenizer.
To store statements in a LM, we train until the model reaches 99 percent memorization accuracy, i.e., overfits the training data almost perfectly, or stop early if accuracy does not improve for 20 epochs.
See \appx\ref{sec:training-details} for training details.

\subsection{Symbolic Representation}

\begin{figure}[!t]
	\centering
	\includegraphics[width=\linewidth,keepaspectratio]{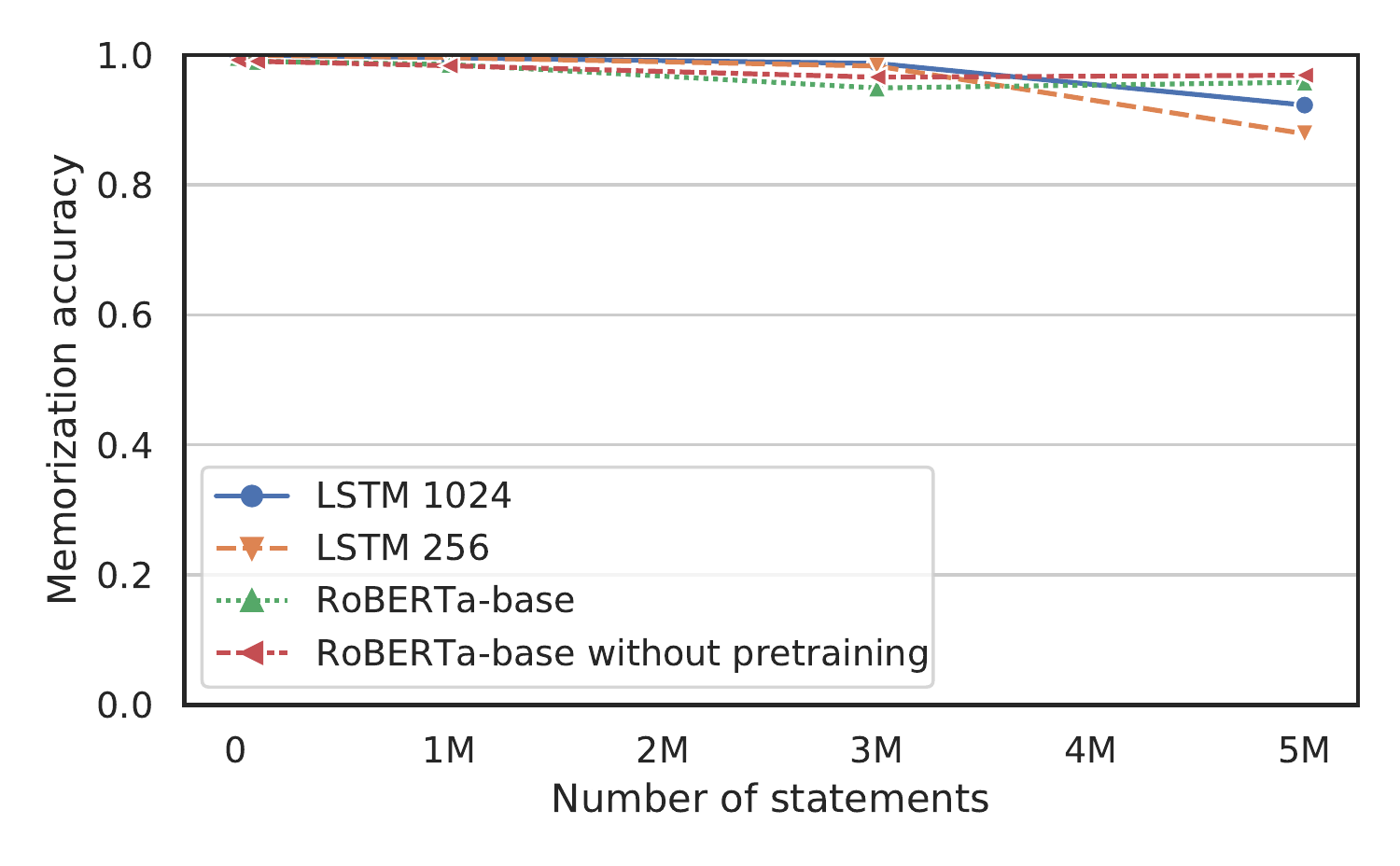}
	\vspace{-4ex}
	\caption{Accuracy of statement memorization with symbolic representation of 1 million entities.}
	\label{fig:symbol-results}
\end{figure}

With symbolic representation, each entity is represented as an entry in the LM's vocabulary.
Prediction is done via masked language modeling \cite{devlin2019bert}, by encoding the query with the LM, projecting the final hidden state of the \mask{} token onto the vocabulary and then taking a softmax over the vocabulary.
As the results show (\fig\ref{fig:symbol-results}), symbolic representation yields very high memorization accuracies with a vocabulary of 1 million entities.
Randomly initialized RoBERTa-base without pretraining works best, memorizing 97 percent of 5 million statements correctly.

Unfortunately, the softmax computation becomes prohibitively slow as the vocabulary size increases \citep{morin2005hierarchical}, making symbolic representation with a softmax over a vocabulary consisting of the full set of 6 million Wikipedia entities impractical.
Imposing a hierarchy is a common approach for dealing with large vocabularies, but did not work well in this case (See \appx\ref{sec:hierarchy}).

\subsection{Surface Form Representation}

With surface form representation, each entity is represented by its canonical name.\footnote{We use English Wikidata labels as canonical names.}
Since this name generally consists of more than one token, we cast memorizing statements and querying facts as a sequence-to-sequence task \cite{sutskever2014sequence}: Given the source sequence \ex{Bert is a character on \mask}, the model should generate the target sequence \ex{Sesame Street}.\footnote{The \mask{} token is included since the target entity does not always occur at the end of a statement.}
To make models memorize statements, we train until perplexity on the training data reaches 1.0 or does not improve for 20 epochs.
For evaluation, we generate the target sequence -- i.e., the answer to a given query -- via a beam search with beam size 10.
We measure perfect-match accuracy of the full entity name, i.e., there is no credit for partial token matches.

The four models under comparison are now treated as sequence-to-sequence encoders and extended with a decoder of the same size: LSTM decoders for LSTM encoders (\emph{LSTM2LSTM}) and randomly initialized Transformers for Transformer encoders (\emph{RoBERTa2Transformer}, \emph{Transformer2Transformer}).

\begin{figure}[!t]
	\centering
	\includegraphics[width=\linewidth,keepaspectratio]{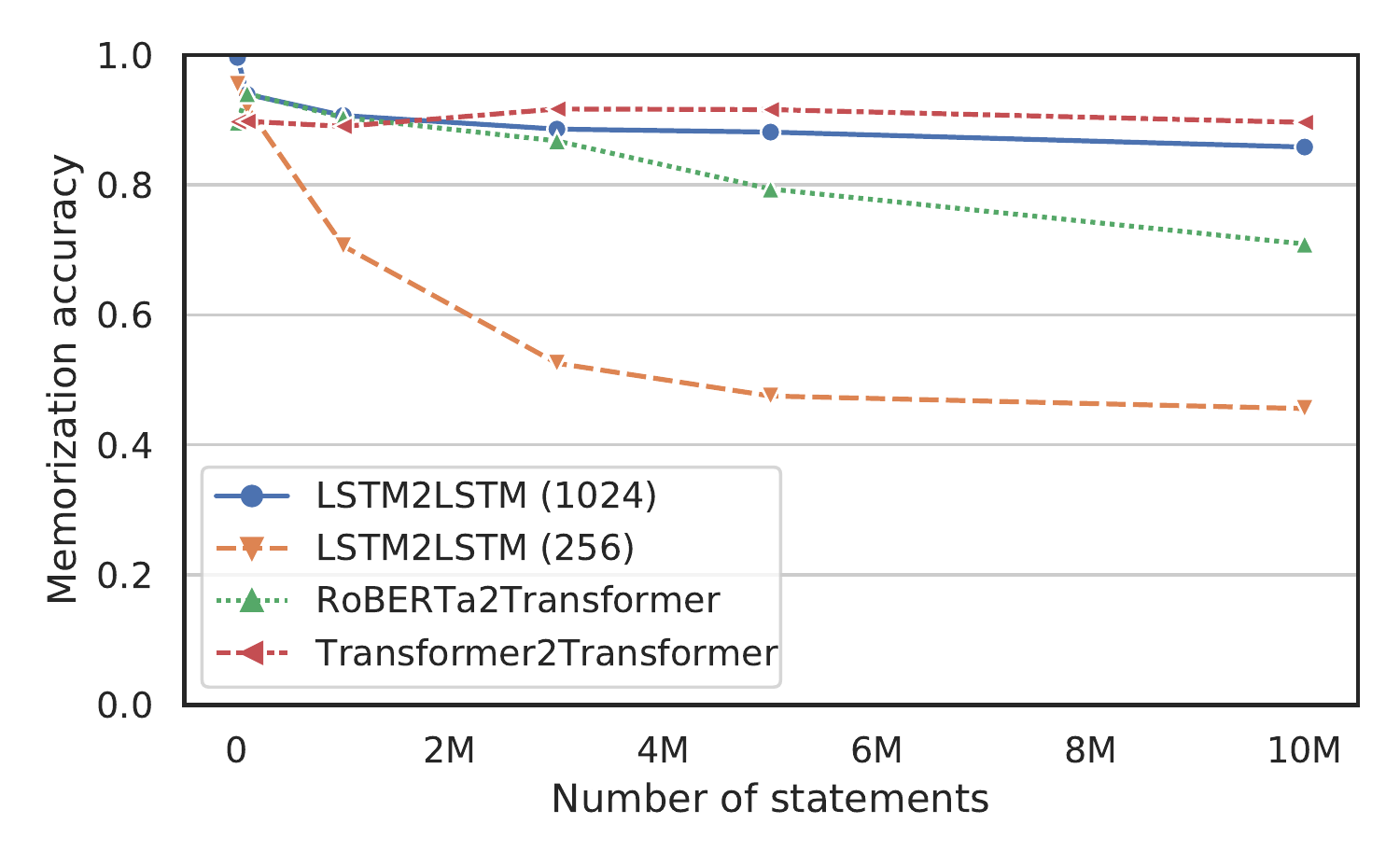}
	\vspace{-3ex}
	\caption{Accuracy of statement memorization with object entities represented by surface forms.}
	\label{fig:surface-results}
\end{figure}

Unlike symbolic representation, surface representation can handle the entire set of 6 million Wikipedia entities.
As with symbolic representation, the randomly initialized Transformer (\fig\ref{fig:surface-results}, dash-dotted red line) has the highest capacity, memorizing 10 million statements with 90 percent accuracy.
A pretrained encoder (\emph{RoBERTa2Transformer}) appears to have a deleterious effect, yielding lower accuracies than the randomly initialized \emph{Transformer2Transformer}.
While the larger \emph{LSTM2LSTM} (layer size 1024) almost matches the performance of the best Transformer model, the smaller one (layer size 256) has insufficient capacity, memorizing less than 50 percent of 5 million statements.

\begin{figure}[!t]
	\centering
	\vspace{1ex}
	\includegraphics[width=\linewidth,keepaspectratio]{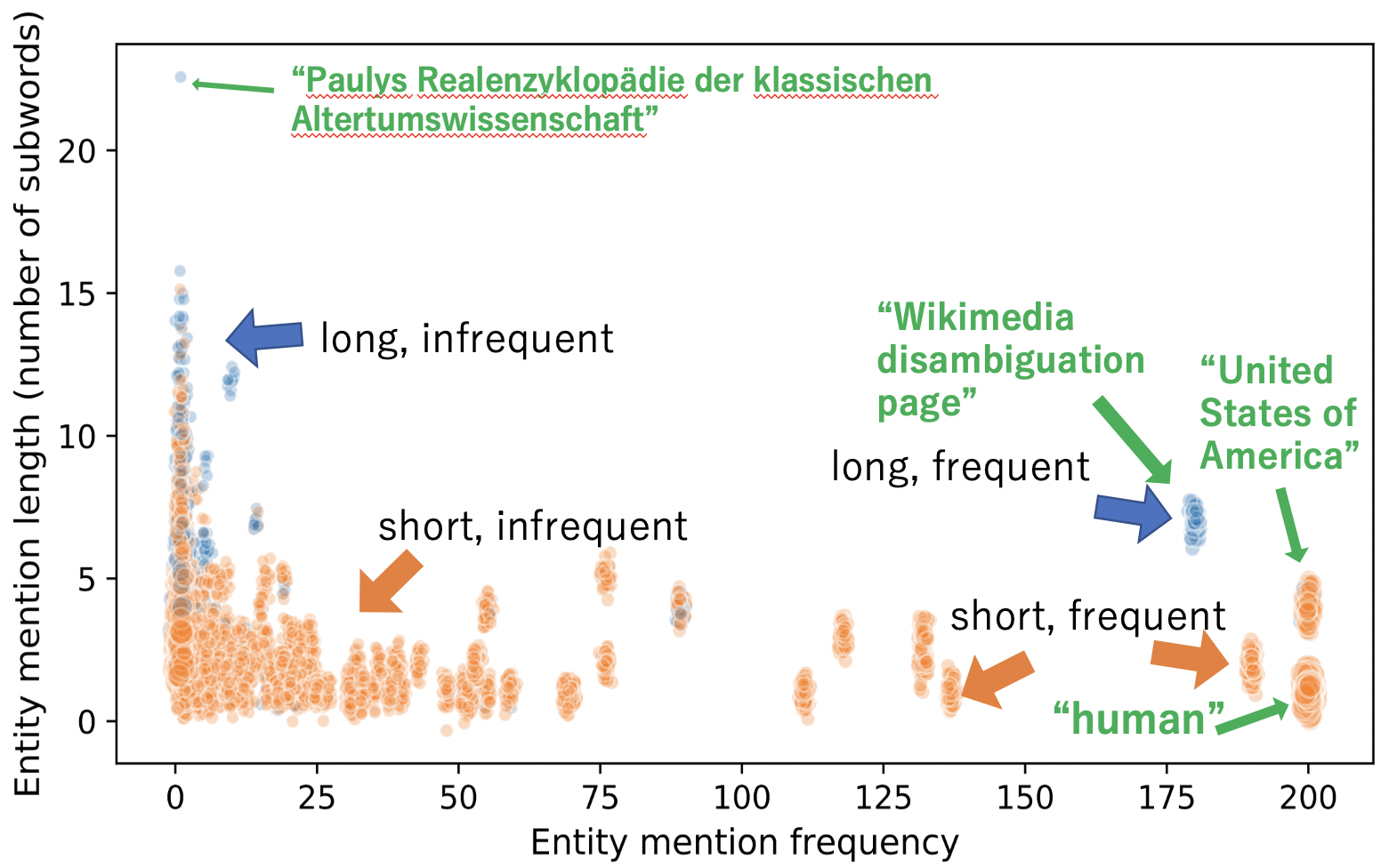}
	\vspace{-2ex}
	\caption{%
		Error analysis of statements memorized via surface form representation.
		Correctly memorized statements orange, wrong ones blue.
		Selected clusters are annotated with the statement's object entity (green).
		Frequencies clipped to 200, jitter applied for clarity.}
	\label{fig:surface-errors}
\end{figure}

Analysis of the Transformer2Transformer model (\fig\ref{fig:surface-errors}) reveals, perhaps unsurprisingly, that statements involving infrequent, long entity mentions are difficult to memorize.%
\footnote{We speculate that this drawback can be mitigated by shortening canonical names while ensuring a one-to-one mapping to entities, but leave this to future work.}
For example, the model fails to memorize most entity mentions that occur only in one to ten statements and have a length of 12 or more subwords (blue cluster, upper left).

\subsection{Continuous Representation}

With continuous representation, an entity $e_i, i \in \interval{1}{N_{entities}}$ is represented by a $d$-dimensional embedding $\mathbf{y}_i \in \mathbb{R}^d$.
After encoding a query with the LM, prediction is performed by projecting the final hidden state corresponding to the \mask{} token onto $\mathbb{R}^d$, obtaining the predicted embedding $\mathbf{\hat{y}} \in \mathbb{R}^d$.
We use fixed, pretrained entity embeddings and train with cosine loss $L = 1 - \cos(\mathbf{\hat{y}}, \mathbf{y_i})$.
At test time, the model prediction $\mathbf{\hat{y}}$ is mapped to the closest pretrained entity embedding $\mathbf{y}_i$ via nearest-neighbor search \citep{johnson2017faiss}.

\begin{figure}[!t]
	\centering
	\includegraphics[width=\linewidth,keepaspectratio]{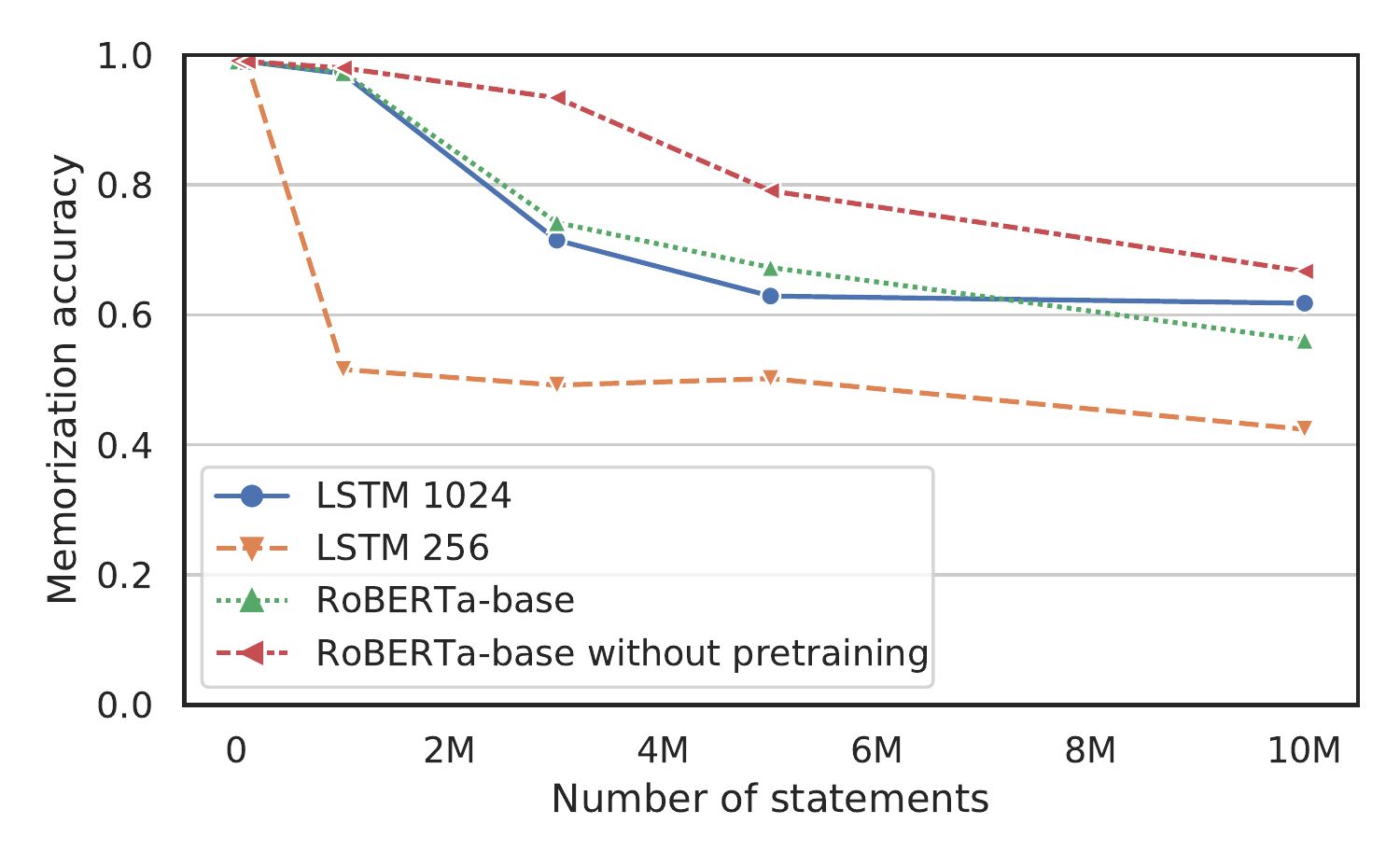}
	\vspace{-3ex}
	\caption{Accuracy of statement memorization with continuous entity representation.}
	\label{fig:embedding-results}
\end{figure}

\begin{figure}[!t]
	\centering
	\vspace{1ex}
	\includegraphics[width=.88\linewidth,keepaspectratio]{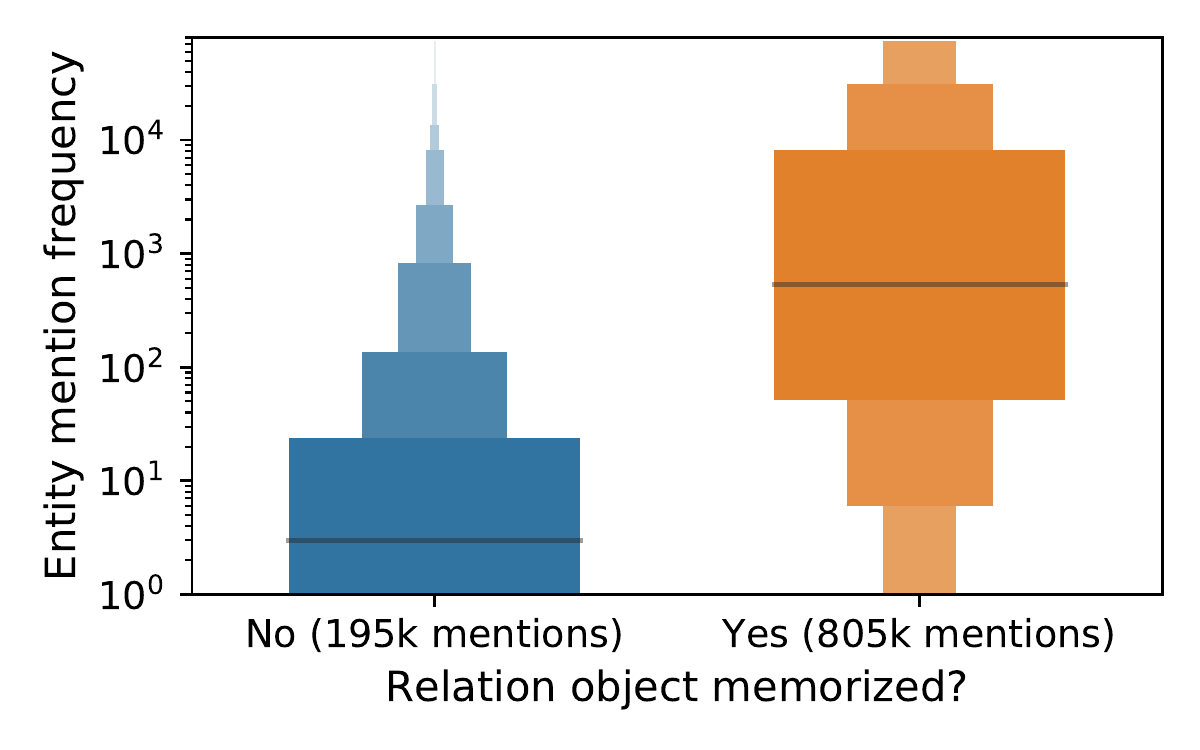}
	\vspace{-1ex}
	\caption{Error analysis of a sample of 1 million statements memorized by a randomly initialized Transformer with continuous representation.}
	\label{fig:embedding-errors}
\end{figure}

\inparatitle{Continuous prediction with fixed, pretrained embeddings} When training randomly initialized embeddings with a similarity objective, a degenerate solution is to make all embeddings the same, e.g., all-zero vectors.
To prevent this, it is common practice to use negative samples \citep{bordes2013translating}.
When using fixed, pretrained embeddings as supervision signal, negative sampling is not necessary, since the target embeddings are not updated and therefore cannot become degenerate.

\inparatitle{Wikidata embeddings} We train embeddings for 6 million Wikidata entities using feature-specific autoencoders to encode entity features such as names, aliases, description, entity types, and numeric attributes, following prior work on multi-modal KB embeddings \citep{pezeshkpour2018embedding} and KB embeddings with autoencoders \citep{takahashi2018interpretable}.
Embedding training is detailed in \appx\ref{sec:embeddings}.

\inparatitle{Results} \fig\ref{fig:embedding-results} shows memorization accuracies achieved with continuous representation.
Like surface representation, continuous representation scales to 6 million entities, and we see the same relative order of models, but with overall lower accuracies.
RoBERTa without pretraining has the highest capacity for storing world knowledge statements, memorizing 67 percent of 10 million statements, while the small LSTM 256 model has the lowest capacity, memorizing 42 percent.
Although far from fully understood, sequence-to-sequence architectures are relatively mature, with highly-optimized toolkits and hyperparameter settings publicly available \citep{ott2019fairseq}.
In contrast, prediction of continuous representations is still in an early stage of research \citep{kumar2019vmf}.
We therefore see these results as lower bounds for LM capacity with continuous representations.

By design, memorization with continuous representations does not rely on entity names, and hence, in contrast to surface form representation, does not lead to difficulties in handling entities with long names.
However, as with surface form representation, infrequent entities are more difficult to memorize than frequent ones.
Most of the memorization errors (\fig\ref{fig:embedding-errors}, blue, left) involve infrequent entities with a median frequency of 3, while most of the correctly memorized statements (orange, right) involve entities that occur more than 100 times.

\section{LM Capacity for Storing Facts}
\label{sec:capacity}

\begin{figure}[t!]
	\centering
	\includegraphics[width=\linewidth]{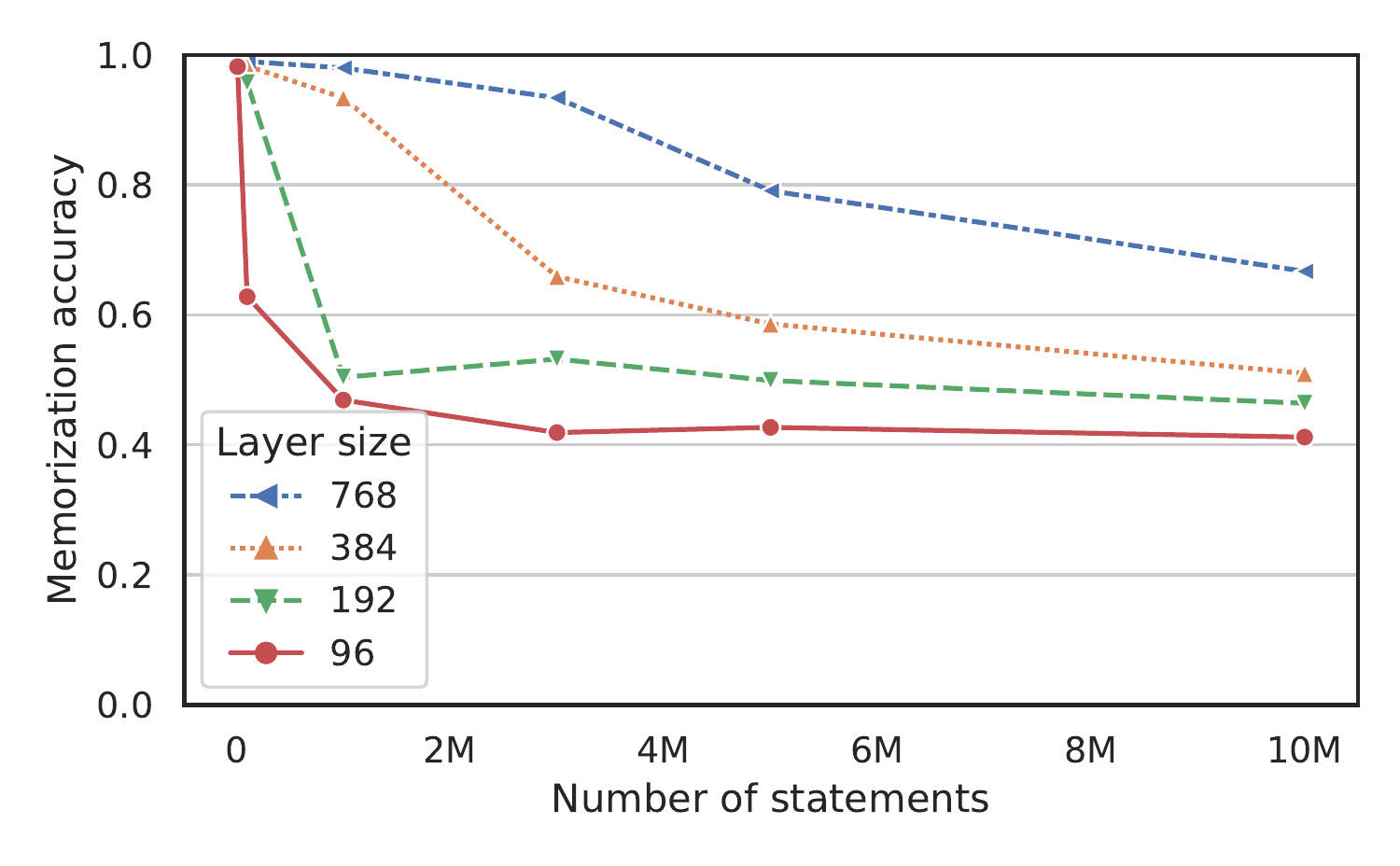}

	\vspace{1ex}

	\includegraphics[width=\linewidth]{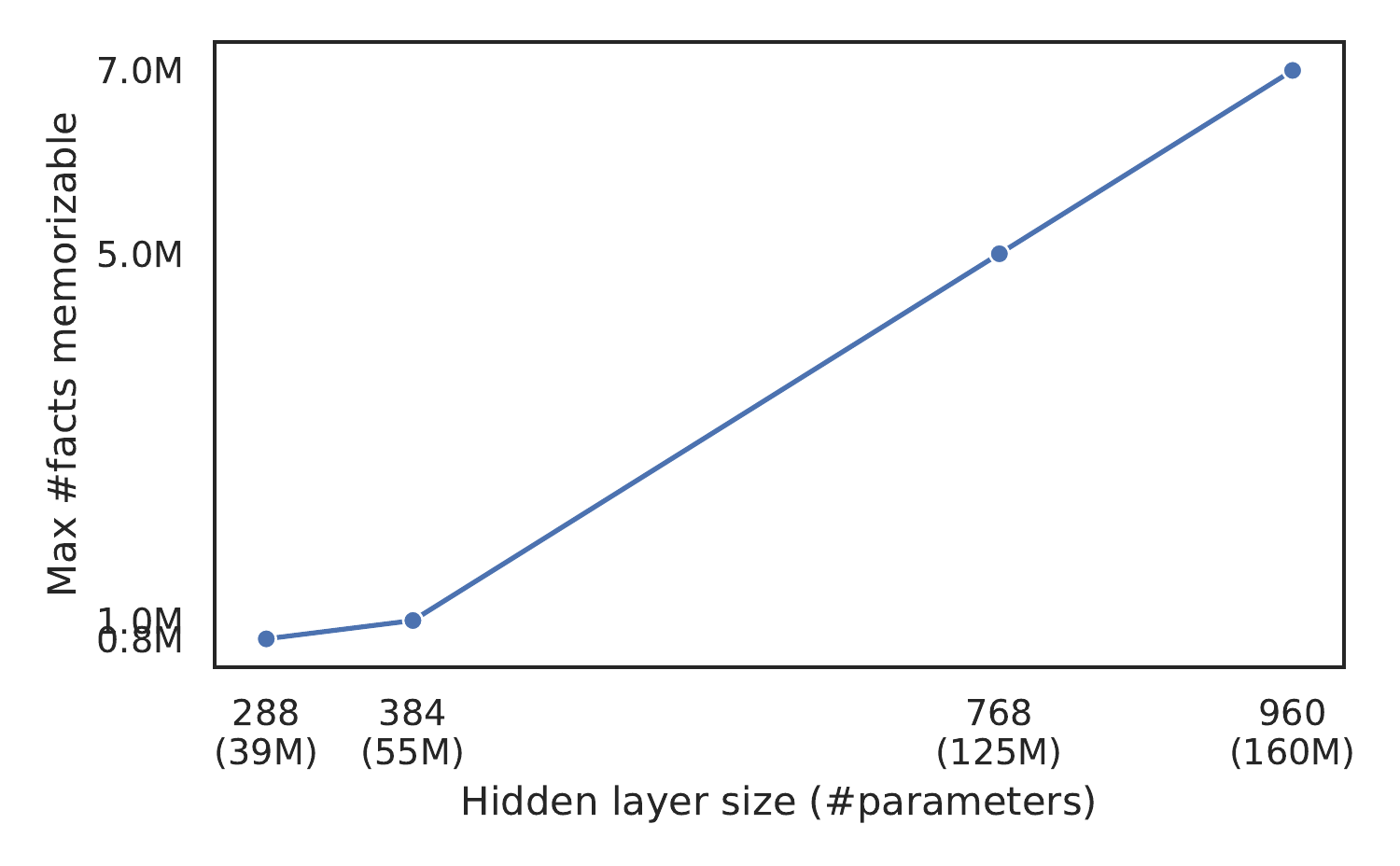}
	\vspace{-4ex}
	\caption{Scaling of storage capacity with model size.
	Memorization accuracy decreases as the number of facts grows (top).
	The maximum number of facts a model of a given layer size (parameter count) can memorize with an accuracy of 80 percent increases linearly (bottom).
	All models are 12-layer Transformer with continuous representation of 6 million entities.}
	\label{fig:scaling}
\end{figure}

We now turn to the second question, how model capacity scales with model size (\fig\ref{fig:scaling}, top).
With a 12-layer Transformer of layer size 96 or 192 (top subfigure, solid red and dashed green lines), memorization accuracy quickly drops as the number of facts to memorize increases.
Larger models can memorize more facts, but accuracy drops rather quickly, e.g., to 65 percent of 3 million facts memorized with a layer size of 384 (dotted orange line).

Assuming a desired memorization accuracy of 80 percent, we record the maximum number of facts a model of a given size can memorize at this level (\fig\ref{fig:scaling}, bottom).
For the model sizes considered here, storage capacity appears to scale linearly, with a model of layer size 384 (55M parameters) storing one million facts and a model of layer size 960 (160M parameters) storing 7 million facts.

Apart from the number of facts to store, we hypothesize that successful storage depends on two more factors: the number of entities and the entropy of their distribution.
As expected, a large number of entities makes memorization more difficult (Table~\ref{tbl:vocab-size}).
The number of entities has a small effect with surface representation (2 percent drop), but with continuous representation accuracy drops from 85 percent to 79 percent when the number of entities increases from 1 to 6 million.
We also observe an impact of the entity distribution (\appx\ref{sec:graph-types}), but leave detailed analysis to future work.

\begin{table}[t!]
	\vspace{1ex}
	\centering
	\adjustbox{max width=\linewidth}{
	\small
	\begin{tabular}{lrr}
		\toprule
		 & \multicolumn{2}{c}{Accuracy} \\
		Representation & 1M & 6M \\
		\midrule
		Symbolic & 0.97 & n/a \\
		Surface & 0.92 & 0.90 \\
		Continuous & 0.85 & 0.79 \\
		\bottomrule
	\end{tabular}
	}
	\caption{The number of entities (1M or 6M) impacts memorization accuracy. The model is a 12-layer Transformer, layer size 768, memorizing 1 million facts.}
	\label{tbl:vocab-size}
\end{table}

\subsection{Storage Efficiency}
\begin{figure}[t!]
	\centering
	\includegraphics[width=\linewidth]{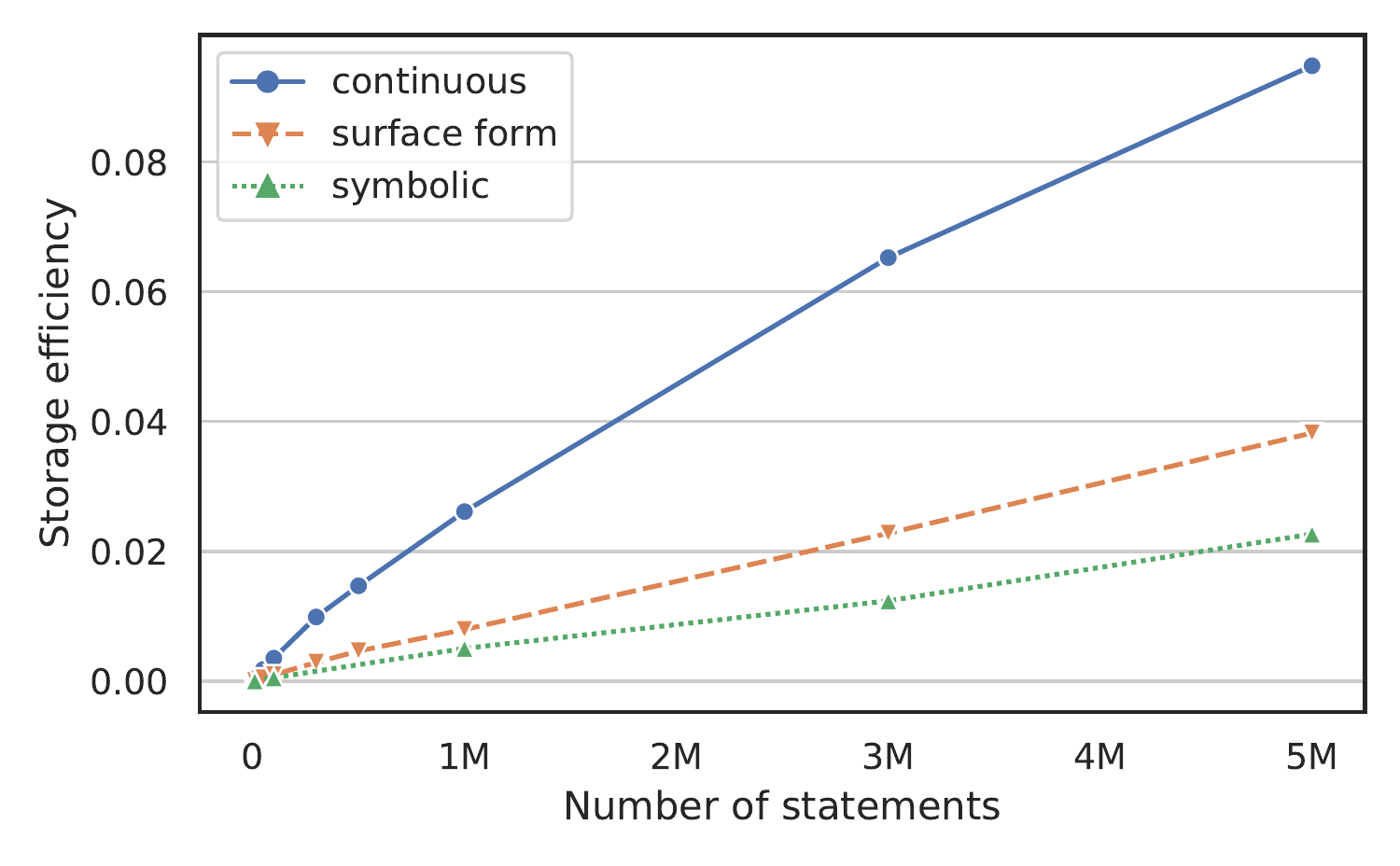}
	\caption{%
		Storage efficiency with symbolic, surface form, and continuous representation of entities.
		In the setting considered in this work, continuous representation is most efficient.
	}
	\label{fig:efficiency}
\end{figure}

Our comparison of different entity representations \secref{sec:entity-representations} does not control for the number of trainable model parameters.
That is, we selected common architectures, such as a Transformer with $12$ layers of size $768$, but made no effort to ensure that, e.g., the number of trainable parameters introduced by the softmax layer in a model with symbolic representation matches the number of trainable parameters introduced by the addition of a sequence-to-sequence decoder component in a model with surface form representation.
In order to more fairly compare entity representations across models with differing numbers of trainable parameters, we formulate the \emph{storage efficiency} of a model designed to memorize statements:
$$\textrm{Storage efficiency} = \frac{\mathit{\#statements} \times \mathit{accuracy}}{\#parameters}$$
This measure expresses the intuition that a model is efficient if it requires few parameters to memorize a large number of statements with high accuracy.
When quantifying efficiency with this measure, we find that continuous representation is the most efficient (Figure~\ref{fig:efficiency}) and hence use this form of entity representation in the remainder of this work.

\label{sec:efficiency}

\section{Querying Stored Facts}
\label{sec:query}

So far, we saw that it is possible to store millions of facts in a LM, by finetuning the model to predict the masked object of statements like \emph{Barack Obama was born in \mask .}
However, given the large number of model parameters and training effort, mere storage is not a compelling achievement: The underlying relations, here \triple{Barack\_Obama}{wasBornIn}{Hawaii}, can be stored more compactly and with perfect accuracy in a structured KB.

One of the potential benefits of the LM-as-KB paradigm is the LM's ability to handle paraphrases.
If the LM's representation of the statement above is sufficiently similar to its representation of queries like \emph{Barack Obama is from \mask} or \emph{Where is Barack Obama from? \mask}, this similarity could allow transfer from the memorized statement to these unseen queries.
Is this soft querying of facts stored in a LM possible?
We now conduct a controlled experiment to answer this question, expecting one of the following three outcomes:

\inparatitle{1.\,Rote memorization} The model memorizes statements with little or no abstraction, so that even small, meaning-preserving changes to the query prevent the model from recalling the correct object.

\inparatitle{2.\,Generic association} The model memorizes pairs of subject and object entities but disregards the predicate.
For example, a model might predict \entity{Hawaii} whenever the query contains the phrase \emph{Barack Obama}, regardless of context.
This pathological behaviour could be especially prevalent if the distribution of object entities co-occurring with a given subject is dominated by one object.

\inparatitle{3.\,Fact memorization} The model memorizes facts expressed in statements by forming abstractions corresponding to entities and predicates.
This allows retrieving a fact with a variety of queries.

Sections~\ref{sec:entity-representations} and \ref{sec:capacity} already established that a model of sufficient size can perform rote memorization of millions of statements.
We now design an experiment to test whether LMs are capable of fact memorization while taking care to distinguish this capability from generic association.
Concretely, our goal is to test if a LM that has memorized a statement like \emph{Barack Obama was born in Hawaii} can use this knowledge to answer a query like \emph{Barack Obama is from \mask}.
Conveniently, \predicate{wasBornIn} relations are among the most frequent in Wikidata and hold for a diverse set of subject and object entities.
This diversity of entities makes this predicate a good candidate for our case study, since statements involving a predicate with a less diverse set of subject or object entities are easier to memorize.\footnote{For example, with the predicate \predicate{isA} and relations like \triple{Barack\_Obama}{isA}{human} the model would do well by always predicting \entity{human} if the subject mention matches a frequent person name pattern such as ``two capitalized words''.}

\begin{figure*}[!t]
	\centering
	\includegraphics[width=\textwidth,height=\textheight,keepaspectratio]{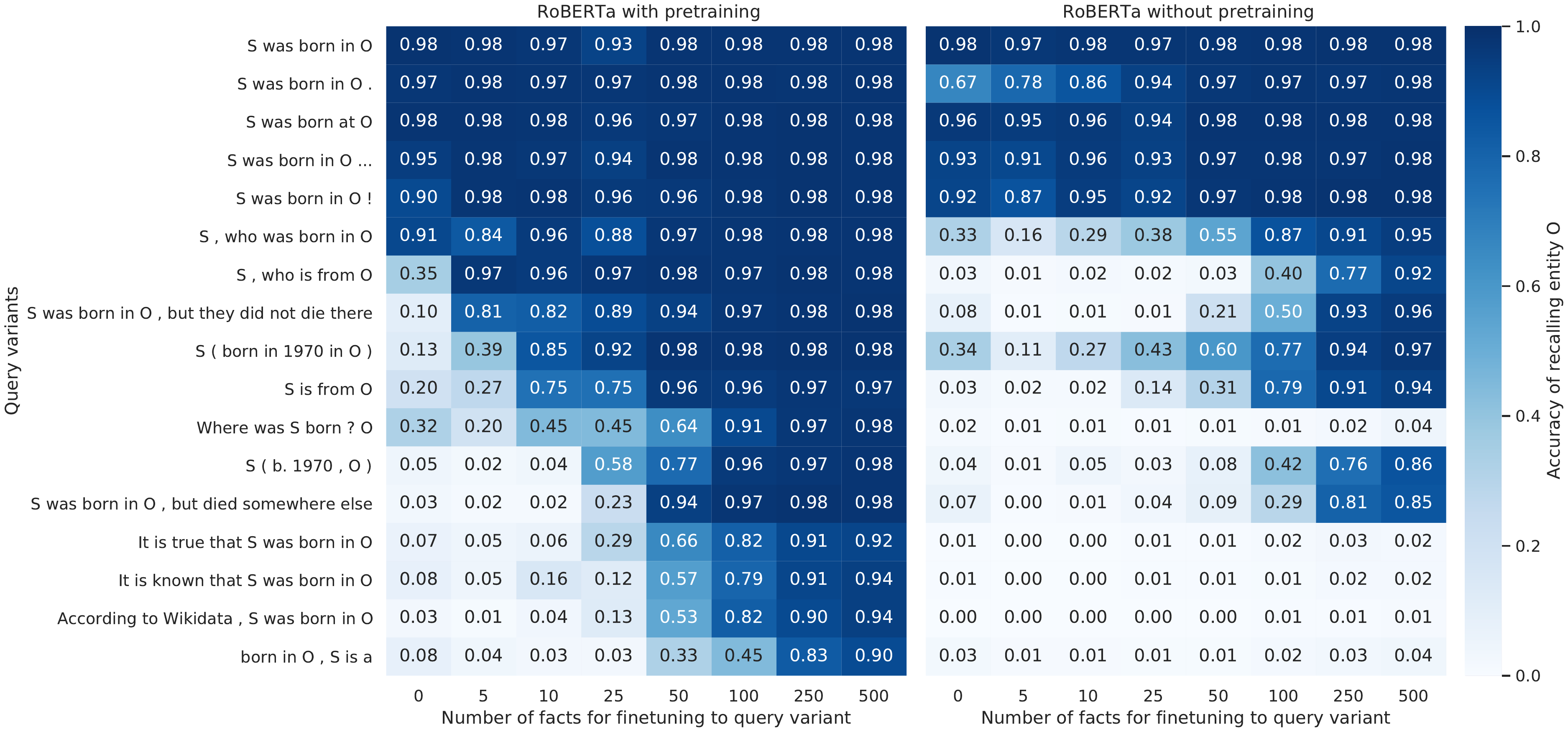}

	\vspace{4ex}

	\includegraphics[width=\textwidth,height=\textheight,keepaspectratio]{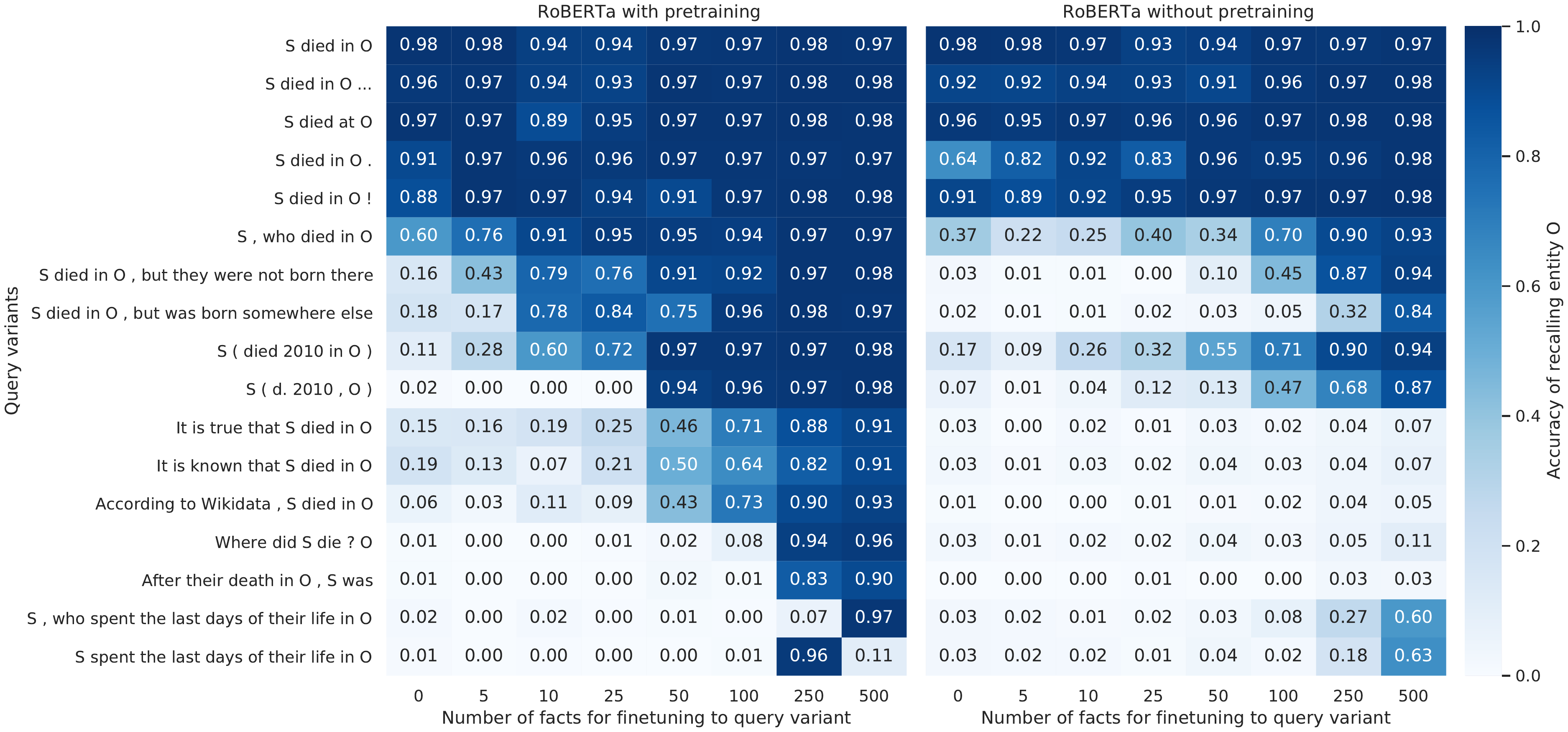}
	\vspace{-1.5ex}
	\caption{Transfer from memorized statements (top: \ex{S was born in O}, bottom: \ex{S died in O}) to query variants.}
	\label{fig:paraphrase-heatmap}
\end{figure*}

\inparatitle{Statements and controls} We sample 100k statements generated by the template \ex{S was born in O}.
To allow distinguishing if a model that memorizes these 100k facts does so by generic assocation or by fact memorization, we introduce \emph{control} facts.
Given a fact \triple{S}{P}{O}, its control \triple{S}{P'}{O'} involves the same subject S, but a distinct predicate P' and object O'.
For example, a control for the fact \triple{Albert\_Einstein}{wasBornIn}{Ulm} is the fact \triple{Albert\_Einstein}{diedIn}{Princeton}.
We add 100k control statements generated from the template \ex{S died in O'} and train RoBERTa-base to memorize all 200k statements with 98 percent accuracy.
The combination of statements and controls counters generic association: To correctly answer the query \ex{Albert Einstein died in \mask}, the model needs to take into account the predicate, since two distinct objects are associated with \entity{Albert\_Einstein}.

\inparatitle{Query variants} Next, we collect query variants, such as \ex{S is from O} (row labels in \fig\ref{fig:paraphrase-heatmap}, top).
Expecting good transfer for variants that are similar to the original statement, we include variants with small changes, such as varying punctuation.
As more diverse variants, we select frequent relation patterns, such as \ex{S (b.\ 1970, O)}, from the GoogleRE Corpus \citep{google2013relation}, as well as a query in question form and queries with irrelevant or misleading distractors such as \ex{S was born in O, but died somewhere else}.
For each variant, we generate 100k queries by filling the S and O slots with the same entity pairs as the original statements. 
To balance statements and controls, we create control templates (row labels in \fig\ref{fig:paraphrase-heatmap}, bottom) and generate a matching number of control statements.

\inparatitle{Transfer results}
We evaluate knowledge transfer from memorized statements to query variants using RoBERTa-base (\fig\ref{fig:paraphrase-heatmap}, top, left), measuring accuracy over the 100k statements generated with a target query variant template.
To measure the effect of pretraining on transfer ability, we compare to RoBERTa-base without pretraining (\fig\ref{fig:paraphrase-heatmap}, top, right).
We consider zero-shot transfer without any finetuning towards the target query variant, and a finetuning setting, in which the LM is first trained to memorize all 100k original statements and then finetuned until it memorizes a small number of statements in the target query format.%
\footnote{Our experiments, which test whether a LM can transfer a memorized fact to given target paraphrases, are converse to the probing setup by \citet{jiang2020how}, which aims to find the best paraphrase for querying a given fact from a LM.}

 In the zero-shot setting (leftmost column),  even small changes to the query lead to a drop in fact recall: Adding an ellipsis (4th row) causes the model to answer 95\% of queries correctly, a 3\% drop from 98\% memorization of the original statements (first row).
 Adding an exclamation mark (5th row) results in a 8\% drop.
 For other paraphrases, e.g., \emph{S, who is from O} (7th row) and \emph{S is from O}, zero-shot transfer works only in 35\% and 20\% of cases, and the question format (11th row) allows zero-shot transfer with 32\% accuracy.
For the remaining paraphrases, e.g., those with parentheticals or the distractor \emph{died}, zero-shot transfer is poor, with accuracies ranging from 3\% to 13\%.

A clear trend is visible: transfer works best for similar statements and worst for dissimilar ones.
To quantify this trend, we compute a representation of a statement template by averaging over its 100k mean-pooled, LM-encoded statements, and then measure the Euclidean distance between the original template representation and the representation of a query variant template.
Correlating Euclidean distance and accuracy of zero-shot transfer obtains a Pearson coefficient of $-0.68$, indicating a strong negative correlation.
That is, transfer tends to work well for paraphrased queries the LM deems similar to the originally memorized statement, but fails if the LM's representation of a query is too dissimilar to its representation of the original statement.
This trend is also reflected in the finetuning setting, with less similar variants requiring up to 500 instances until the model achieves 90 percent accuracy (last row), while transfer to more similar variants works well after finetuning on 5 to 50 target instances.

When using RoBERTa without pretraining to memorize statements, knowledge transfer to query variants is much worse.
While transfer still works for the most similar variants (right, top rows), less similar variants require more finetuning compared to pretrained RoBERTa (right, middle rows).
Transfer does not work for the least similar variants, with accuracies as low as 1 to 4 percent even after finetuning with 500 instances (right, bottom rows).
Similar results for control statements are shown in \fig\ref{fig:paraphrase-heatmap} (bottom).
We take these results as evidence that pretraining enables LMs to handle paraphrased queries and that LMs can memorize facts beyond mere rote memorization and generic association.

\section{Limitations and Conclusions}

\inparatitle{Limitations} This work is not without limitations.
We only use one KB in our experiments.
Arguably, as the largest publicly available source of world knowledge, Wikidata is the most promising resource for equipping LMs with such knowledge, but attempts to store a KB with different structure might result in different outcomes, since some types of graphs are easier to memorize for a LM than others (See \appx\ref{sec:graph-types}).

While we use language like \ex{\emph{train} a LM to memorize statements} for simplicity throughout this work, what we do in case of pretrained LMs is more akin to adaptive pretraining \citep{gururangan2020dont}.
It is possible that integrating entity supervision directly into LM pretraining \citep{fevry2020entities} allows more efficient fact storage.

Our analysis was focused on entity representations and ignored the question of how to represent relation predicates or entire relation triples.
Here, relation learning \citep{baldini2019matching} and LM pretraining on fact-aligned corpora \cite{elsahar2018trex} are avenues for future work.

Finally, we formulated the LM-as-KB paradigm in terms of storing and retrieving relation triples.
While structured KBs such as Wikidata consist of such triples and hence our experiments showing storage and retrieval of triples LMs are sufficient as a proof-of-concept in principle, structured KBs allow more complex queries than the ones considered here, such as 1-to-n relations, multihop inference, queries involving numerical ranges, or facts qualified by time and location \citep{hoffart2013yago2}.

\inparatitle{Conclusions} We gave a positive answer to \citet{petroni2019language}`s question if language models can serve as knowledge bases.
Arguing that treating LMs as KBs requires representing a large number of entities, storing a large number of facts, and the ability to query a fact with a variety of queries, we showed that current LM architectures fulfill these requirements when extended with a component for representing entities.
In addition to the ability to handle paraphrased queries, we envision further benefits from the LM-as-KB paradigm.
For example, the fact-memorization and paraphrase-finetuning setting introduced in Section~\ref{sec:query} allows precise control over which facts a LM learns.

\section{Acknowledgments}
We thank the anonymous reviewers for helpful feedback.
This work was supported by a Google Focused Research Award.

\bibliography{lit}
\bibliographystyle{acl_natbib}

\onecolumn
\newpage
\appendix

\section{Overview: world knowledge in natural language processing}
\label{sec:relwork-tbl}

\begin{table*}[h!]
	\adjustbox{max width=\linewidth}{
	\small
	\begin{tabular}{p{.2\linewidth} | p{.15\linewidth}p{.15\linewidth} | p{.5\linewidth}}
		\toprule
		Paradigm / Task & Input & Output & Models and objectives \\
		\midrule
		Language modeling & Text & Text & Next word prediction \citep{shannon1948mathematical,elman1990finding,bengio2003neural}, masked token prediction \citep{devlin2019bert} \\
		LM-as-KB? & Text & Text / single-token entity name & Closed-book QA \citep[LAMA probe,][]{petroni2019language} \\
		Sequence-to-sequence & Text & Text & Text-to-text transformer \citep[T5,][]{raffel2019exploring}, closed-book QA \citep{roberts2020knowledge} \\
		Retrieval & Text & Text, answer span & Answer-span selection \citep{chen2017reading}, retrieval-augmented LM \citep{guu2020realm}, open-book QA \\
		\midrule
		Entity replacement & Text, entity mention spans & Text & Detecting replaced entity mentions \citep{xiong2019pretrained} \\
		Entity linking (EL) & Text, entity mention spans & Target entity & AIDA \citep{hoffart2011robust}, neural EL \citep{francislandau2016capturing,kolitsas2018end} \\
		Entity embeddings & Text, entity mention spans & Entity embeddings & Joint embedding of entities and text \citep{yamada2016joint} \\
		LM with entity embeddings & Text, linked entity mentions, entity embeddings & Text & ERNIE \citep{zhang2019ernie}, E-BERT \citep{poerner2019ebert} \\
		LM with integrated EL & Text, entity embeddings & Text & KnowBert \citep{peters2019knowledge} \\
		\textbf{LM-as-KB (this work)} & Natural language query & Target entity & Fact memorization, paraphrased queries, closed-book QA \\
		Knowledge-aware LM & Text, knowledge (sub)graph & Target entity, text & Neural Knowledge LM \citep{ahn2016neural}, Reference-aware LM \citep{yang2017reference}, Knowledge graph LM \citep{logan2019baracks} \\
		Semantic parsing & natural language query & meaning representation, target entity & SEMPRE \citep{berant2013semantic}, GNNs for KBQA \citep{sorokin2018semantics} \\
		Universal Schema & relation triples, text patterns & entity tuple and relation embeddings & Matrix factorization \citep{riedel2013relation} \\
		\midrule
		Knowledge graph embeddings & relation triples & node and edge embeddings & Link prediction; RESCAL \citep{nickel2011three}, TransE \citep{bordes2013translating}, ComplexE \citep{trouillon2016complex}, ConvE \citep{dettmers2018conve} \\
		Graph neural networks & nodes, node features, edges & node embeddings & DeepWalk \citep{perozzi2014deepwalk}, graph neural networks \citep{kipf2017gcn} \\
		Knowledge graphs & nodes, edges & nodes, edges & Storage and retrieval, SQL/SPARQL queries, symbolic reasoning \cite{coppens2013reasoning} \\
		\bottomrule
	\end{tabular}
	}
	\caption{Approaches for using world knowledge in natural language processing, ranging from unstructured, purely text-based approaches (top), over approaches that mix text and structured KBs to varying degrees (middle), to approaches operating on structured KBs (bottom).}
	\label{tbl:rel-work}
\end{table*}

\newpage

\onecolumn
\section{Templates for generating English statements from Wikidata relations}
\label{sec:templates}

\begin{table}[h!]
\begin{multicols}{2}
\small
\centering
\tablefirsthead{ID & Template \\
\midrule}
\tablehead{ID & Template \\
\midrule}
	\begin{supertabular}{rl}

		P31 & S is an instance of O \\
		P106 & S has the occupation O \\
		P17 & S belongs to the country O \\
		P131 & S is located in the administrative territorial entity O \\
		P27 & S is citizen of O \\
		P47 & S shares a border with O \\
		P19 & S was born in O \\
		P161 & S has the cast member O \\
		P421 & S is located in time zone O \\
		P166 & S received the award O \\
		P54 & S is a member of the sports team O \\
		P20 & S died in O \\
		P136 & S has the genre O \\
		P69 & S was educated at O \\
		P1412 & S is a language spoken, written or signed in O \\
		P190 & S is a twinned administrative body of O \\
		P641 & S participates in the sport O \\
		P150 & S contains the administrative territorial entity O \\
		P463 & S is a member of O \\
		P735 & S has the given name O \\
		P1343 & S is described by source O \\
		P361 & S is a part of O \\
		P159 & the headquarters of S are located in O \\
		P1344 & S is participant of O \\
		P495 & S has the country of origin O \\
		P39 & S held the position of O \\
		P910 & S has the main category O \\
		P105 & S has the taxon rank O \\
		P527 & S has the part O \\
		P108 & S is employed by O \\
		P279 & S is a subclass of O \\
		P171 & S has the parent taxon O \\
		P140 & S has the religion O \\
		P407 & S is in the O language \\
		P1303 & S plays the instrument O \\
		P1411 & S has been nominated for O \\
		P102 & S is a member of political party O \\
		P3373 & S is a sibling of O \\
		P1376 & S is the capital of O \\
		P509 & S died because of O \\
		P937 & S works in O \\
		P264 & S was produced by the record label O \\
		P119 & S is buried in O \\
		P138 & S is named after O \\
		P530 & S has diplomatic relations with O \\
		P40 & S is a child of O \\
		P155 & S follows O \\
		P276 & S is located in O \\
		P156 & S is followed by O \\
		P36 & S has the capital O \\
		P1196 & S has the manner of death O \\
		P127 & S is owned by O \\
		P101 & S works in the field O \\
		P607 & S participated in the conflict O \\
		P364 & S is a film or TV show with the original language O \\
		P6379 & S has works in the collection O \\
		P1346 & S is a winner of the O \\
		P22 & S is the father of O \\
		P137 & S is operated by O \\
		P413 & S plays the position O \\
		P26 & S is spouse of O \\
		P1830 & S is owner of O \\
		P1454 & S has the legal form O \\
		P206 & S is located in or next to body of water O \\
		P710 & S is a participant of O \\
		P1441 & S is present in the work O \\
		P1532 & S represents O when playing sport O \\
		P86 & S was composed by O \\
		P840 & S is set in the location O \\
		P172 & S belongs to the ethnic group O \\
		P175 & S is performed by O \\
		P57 & S is directed by O \\
		P1889 & S is different from O \\
		P162 & S is produced by O \\
		P118 & S belongs to the league O \\
		P58 & S is screenwritten by O \\
		P551 & S has the residence O \\
		P103 & S has the native language O \\
		P2789 & S connects with O \\
		P750 & S has the distributor O \\
		P725 & S is voiced by O \\
		P272 & S is produced by the company O \\
		P112 & S was founded by O \\
		P452 & S belongs to the industrial sector O \\
		P81 & S is connected to line O \\
		P97 & S has noble title O \\
		P740 & S formed in the location O \\
		P360 & S is a list of O \\
		P793 & S is associated with the significant event O \\
		P915 & S was filmed at O \\
		P410 & S has military rank O \\
		P1001 & S applies to the jurisdiction of O \\
		P30 & S is located on the continent O \\
		P749 & S has parent organization O \\
		P1435 & S has heritage designation O \\
		P53 & S belongs to the family of O \\
		P400 & S was developed for the platform O \\
		P921 & S has the main subject O \\
		P37 & S has the official language O \\
		P734 & S has the family name O \\

	\end{supertabular}
\end{multicols}
	\caption{Templates used to generate English statements from Wikidata relations.}
\end{table}

\newpage

\section{Random sample of English statements generated from Wikidata relations}
\label{sec:statements}
{
	\small
	\begin{itemize}
		\item	The Underfall Yard is followed by English Electric Part One
		\item	Gazi Beg is a child of Farrukh Yassar
		\item	2011 European Rowing Championships is followed by 2012 European Rowing Championships
		\item	2009 Yemeni tourist attacks is located in Shibam
		\item	George Best – A Tribute is performed by Peter Corry
		\item	Gamecock Media Group is owned by SouthPeak Games
		\item	2017–18 Sheffield Wednesday F.C. season is followed by 2018–19 Sheffield Wednesday F.C. season
		\item	Nennslingen is located in or next to body of water Anlauter
		\item	2013–14 Xavier Musketeers men's basketball team is followed by 2014–15 Xavier Musketeers men's basketball team
		\item	Shock to the System is a part of Cyberpunk
		\item	1918–19 Ohio Bobcats men's basketball team follows 1917–18 Ohio Bobcats men's basketball team
		\item	Ramya Krishnan has the spouse Krishna Vamsi
		\item	The Cloud Minders follows The Way to Eden
		\item	Curve is followed by Somethingness
		\item	Austin Road is named after John Gardiner Austin
		\item	Dione juno has the parent taxon Dione
		\item	Spirit Bound Flesh is followed by The Wake
		\item	Sidnei da Silva has the given name Sidnei
		\item	In Memoriam is performed by Living Sacrifice
		\item	Tracks and Traces is followed by Live 1974
		\item	Grumman Gulfstream I is operated by Phoenix Air
		\item	Timeline of Quebec history has the part Timeline of Quebec history (1982–present)
		\item	Edwin C. Johnson held the position of Lieutenant Governor of Colorado
		\item	Here Comes the Summer follows Jimmy Jimmy
		\item	In Custody is screenwritten by Anita Desai
		\item	Bertie Charles Forbes is the father of Malcolm Forbes
		\item	The Mambo Kings has the cast member Helena Carroll
		\item	Carnival of Souls has the cast member Art Ellison
		\item	1995–96 Philadelphia Flyers season is followed by 1996–97 Philadelphia Flyers season
		\item	John Harley is the father of Edward Harley, 5th Earl of Oxford and Earl Mortimer
		\item	Jane Fellowes, Baroness Fellowes has the spouse Robert Fellowes, Baron Fellowes
		\item	Francis of Assisi is buried in Basilica of San Francesco d'Assisi
		\item	1990 Maharashtra Legislative Assembly election follows 1985 Maharashtra Legislative Assembly election
		\item	Makabana Airport is named after Makabana
		\item	Calvin Booth was born in Reynoldsburg
		\item	The Telltale Head is followed by Life on the Fast Lane
		\item	Alajos Keserű is a sibling of Ferenc Keserű
		\item	Long An contains the administrative territorial entity Châu Thành
	\end{itemize}
}

\newpage

\section{Hyperparameter settings and replicability statement}
\label{sec:training-details}

\begin{table*}[h!]
	\small
	\centering
	\begin{tabular}{lllr}
		\toprule
			Entity representation & Architecture & Hyper-param. & Value \\
		\midrule
			Symbolic & LSTM & layers & 2 \\
			 &  & hidden size & 256, 1024 \\
			 &  & dropout & 0.0 \\
			 &  & learning rate & 0.001 \\
			 &  & lr-scheduler & plateau \\
			 &  & optimizer & Adam \\
			 &  &  & \\
			 & Transformer & model name & RoBERTa-base \\
			 &  & layers & 12 \\
			 &  & hidden size & 768 \\
			 &  & learning rate & 5e-5 \\
			 &  & lr-scheduler & plateau \\
			 &  & optimizer & Adam \\
			 &  &  & \\
			Surface form & LSTM & layers (enc) & 2 \\
			 &  & hidden size (enc) & 256, 1024 \\
			 &  & layers (dec) & 2 \\
			 &  & hidden size (dec) & 256, 1024 \\
			 &  & learning rate & 0.001 \\
			 &  & lr-scheduler & plateau \\
			 &  & optimizer & Adam \\
			 &  &  & \\
			 & Transformer & model name (enc) & RoBERTa-base \\
			 &  & layers (enc) & 12 \\
			 &  & hidden size (enc) & 768 \\
			 &  & dropout & 0.0 \\
			 &  & model name (dec) & random init. \\
			 &  & layers (dec) & 12 \\
			 &  & hidden size (dec) & 768 \\
			 &  & learning rate & 5e-4 \\
			 &  & lr-scheduler & inverse sqrt \\
			 &  & optimizer & Adam \\
			 &  &  & \\
			Continuous & LSTM & layers & 2 \\
			 &  & hidden size & 256, 1024 \\
			 &  & dropout & 0.0 \\
			 &  & learning rate & 0.001 \\
			 &  & lr-scheduler & plateau \\
			 &  & optimizer & Adam \\
			 &  & entity emb.\, dim & 64 \\
			 &  & entity emb.\, trainable & no \\
			 &  &  & \\
			 & Transformer & model name & RoBERTa-base \\
			 &  & layers & 12 \\
			 &  & hidden size & 768 \\
			 &  & learning rate & 5e-5 \\
			 &  & lr-scheduler & plateau \\
			 &  & optimizer & Adam \\
			 &  & entity emb.\, dim & 64 \\
			 &  & entity emb.\, trainable & no \\

		\bottomrule
	\end{tabular}
	\caption{Hyperparameter settings used in our experiments.}
\end{table*}

\newpage
\section{Embeddings of Wikidata entities}
\label{sec:embeddings}

\begin{figure*}[h!]
	\includegraphics[width=\linewidth]{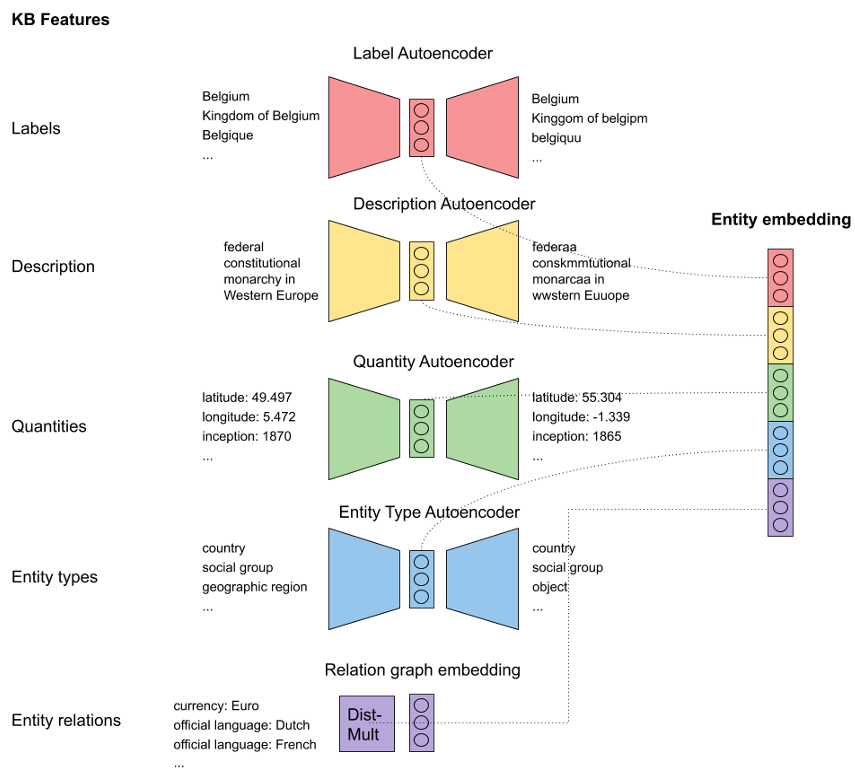}
	\caption{Training embeddings of Wikidata entities with feature-specific autoencoders.}
\end{figure*}

We train the embedding of a given Wikidata entity by collecting its features from, encoding each feature to obtain a dense feature representation, and then concatenating feature representations.
For textual features, we use RoBERTa-base as encoder and train corresponding decoders in a standard sequence-to-sequence auto-encoding setup.
For quantities, we select the 100 most common quantity types to obtain a fixed-sized representation and then follow a standard auto-encoding setup.
Similarly we obtain a fixed-size entity type representation by selecting the 1000 most common entity types.
The concatenated feature-representations are then compressed to embedding size $d$, using a separate autoencoder.
Preliminary experiments with embedding sizes $d \in \left\{64, 128, 192, 256 \right\}$ showed similar memorization accuracies for all $d$, but faster convergence for smaller sizes.
We set $d = 64$ in our main experiments.

\newpage
\twocolumn
\section{Things that didn't work}

\subsection{Hierarchical entity representation with binary codes}
\label{sec:hierarchy}
Since imposing a hierarchy is a common method for dealing with large vocabulary sizes \citep{morin2005hierarchical} in general, and large inventories of entities and entity types in particular \citep{raiman2018deeptype,lopez2019fine}, we created a hierarchy of all entities in Wikidata, using a given entity's position in this hierarchy as training signal. Specifically, we created the entity hierarchy by fitting a KD-tree \citep{bentley1975multidimensional,scipy2020} with leaf size 1 over pretrained entity embeddings, thereby obtaining a binary partitioning of the embedding space in which each final partition contains exactly one entity embedding. The path from the KD-tree's root to a leaf can be represented as a binary code, which we use as training signal \citep{oda2017binary}.
Memorization accuracy of world knowledge facts with object entities represented in the form of these binary codes was substantially lower compared to the three approaches described in the main part of this work.

\subsection{Training entity embeddings with negative sampling}

Instead of using fixed, pretrained entity embeddings as training signal, we experimented with randomly initialized embeddings that are updated during training, using between 1 and 50 in-batch negative samples, which is a standard method in the knowledge base embedding literature \citep{bordes2013translating} and has been used successfully for entity retrieval \citep{gillick2019dense}.
However, compared to using fixed, pretrained entity embeddings without negative sampling, we observed lower memorization accuracies and slower convergence in our experiments.

\subsection{Updating pretrained entity embeddings during training}

Instead of using fixed entity embeddings, we tried updating them during training with in-batch negative sampling.
This increased the number of trainable parameters, memory usage, and training time, but did not lead to higher memorization accuracies.

\subsection{Continuous representation with Euclidean distance loss}

Instead of normalizing entity embeddings to the unit hypersphere and training with cosine loss, we experimented with predicting the original pretrained entity embeddings and using the Euclidean distance as loss.
Compared to using spherical entity embeddings as prediction targets, we observed slower convergence and lower memorization accuracies.

\onecolumn
\newpage

\section{Impact of graph type on memorizability}
\label{sec:graph-types}

\begin{figure*}[h!]
	\includegraphics[width=\linewidth]{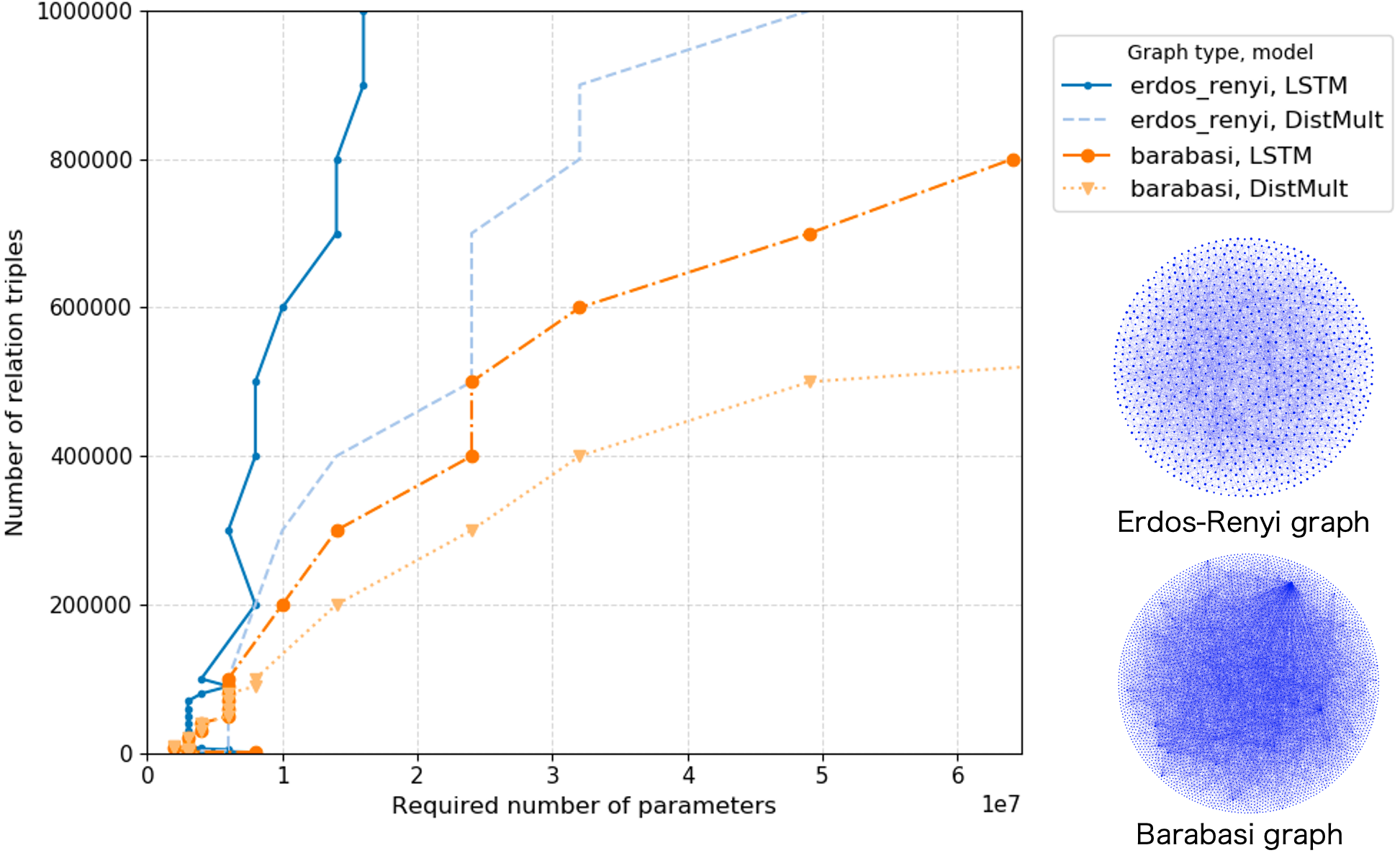}
	\caption{
	Impact of graph type on a model's ability to memorize the graph.
	We consider two types of random graphs, namely a uniform (Erdos-Renyi) graph, and a scale-free (Barabasi) graph.
	We interpret graph edges as relation triples in a knowledge graph and train models to predict the relation object, given subject and predicate, until memorization accuracy reaches 99 percent.
	For a given number of model parameters, we gradually increase the number of relation triples to memorizes and record the maximum number of relation triples memorized for this number of parameters.
	We compare an LSTM, as well as a bilinear KB embedding (DistMult).
	For a given parameter budget, models are able to memorize more triples from a Erdos-Renyi graph (blue) than from a Barabasi graph, indicating that the latter is more difficult to memorize.
	}
\end{figure*}

\end{document}